\documentclass[12pt]{article}
\usepackage[english]{babel}
\usepackage[utf8x]{inputenc}
\usepackage{amsmath}
\usepackage{graphicx}
\usepackage{etoolbox}
\usepackage{changepage}
\usepackage{titlesec}
\usepackage[parfill]{parskip}
\usepackage[margin=1in]{geometry}
\usepackage{times}
\usepackage{float}
\usepackage{lipsum} 

\titleformat*{\section}{\normalsize\bfseries} 

\title{\large \textbf{An Efficient Approach for Muscle Segmentation and 3D Reconstruction Using Keypoint Tracking in MRI Scan}} 
\author{\normalsize Mengyuan Liu, Jeongkyu Lee\\
\normalsize {liu.mengyu, jeo.lee}@northeastern.edu\\
\normalsize Northeastern University, San Jose, USA}

\date{} 

\makeatletter 
\patchcmd{\@maketitle}{\begin{center}}{\begin{adjustwidth}{0.5in}{0.5in}\begin{center}}{}{}
\patchcmd{\@maketitle}{\end{center}}{\end{center}\end{adjustwidth}}{}{}
\makeatother

\begin{document}
\raggedright
\maketitle
\thispagestyle{empty}
\pagestyle{empty}

\section*{Abstract}
Magnetic resonance imaging (MRI) enables non-invasive, high-resolution analysis of muscle structures. However, automated segmentation remains limited by high computational costs, reliance on large training datasets, and reduced accuracy in segmenting smaller muscles. Convolutional neural network (CNN)-based methods, while powerful, often suffer from substantial computational overhead, limited generalizability, and poor interpretability across diverse populations. This study proposes a training-free segmentation approach based on keypoint tracking, which integrates keypoint selection with Lucas-Kanade optical flow. The proposed method achieves a mean Dice similarity coefficient (DSC) ranging from 0.6 to 0.7, depending on the keypoint selection strategy, performing comparably to state-of-the-art CNN-based models while substantially reducing computational demands and enhancing interpretability. This scalable framework presents a robust and explainable alternative for muscle segmentation in clinical and research applications.
\section*{Introduction}
Magnetic resonance imaging (MRI) is an essential tool in medical diagnostics due to its non-invasive and whole-body imaging capabilities. However, the development of techniques to efficiently, and accurately segment individual muscles remains limited. Current methods are mainly based on 2D \cite{zhu2021deep} and 3D \cite{ni2019automatic} convolutional neural networks (CNNs) \cite{ketkar2021convolutional}, which require extensive annotated datasets and significant computational resources. Furthermore, these approaches often struggle with generalizability and underperform in segmenting smaller muscles, with  Dice similarity coefficients (DSCs) \cite{Dice1945measures} ranging from 0.60 to 0.80 \cite{ni2019automatic}. Achieving higher segmentation accuracy often requires retraining the model for different muscles and population groups, further compounding the computational burden. The limited availability of MRI datasets and the labor-intensive nature of annotation further exacerbate these challenges, restricting the adaptability of deep learning models across diverse muscle types and population groups \cite{hesamian2019deep}.

Another critical limitation is the lack of interpretability inherent in CNN-based segmentation \cite{reyes2020interpretability, salahuddin2022transparency}. The outputs generated by these models are often difficult to be explained, particularly when the segmented boundaries are not readily discernible to the human eye.

To address these challenges, we propose a novel approach to muscle segmentation that employs keypoint tracking algorithms to identify anatomical landmarks along muscle boundaries in MRI scans. In contrast to deep learning-based segmentation methods, our approach eliminates the training process. Consequently, high-performance computing, such as GPUs, is not needed for training and testing. Moreover, it offers more interpretable results, facilitating a clearer understanding. This innovation marks a significant advancement toward reducing the computational and data demands of muscle segmentation while maintaining accuracy, generalizability, and interpretability across diverse subject populations.
\section*{Related Work}

This section reviews recent advancements in deep learning methods for medical image segmentation, feature extraction, and tracking. By analyzing existing literature, we emphasize the importance of developing techniques that not only improve segmentation accuracy but also enhance computational efficiency, generalizability, and interpretability.

\textit{Deep Learning Based Segmentation}

Deep learning methods, particularly CNNs, have been widely applied to the task of muscle segmentation in MRI. These models leverage 2D convolutions, 3D convolutions, or hybrid combinations of both to learn hierarchical features that enable accurate delineation of muscle structures \cite{zhu2021deep, ni2019automatic}. Notable architectures, including U-Net \cite{Ronneberger2015u} and 3D U-Net \cite{3dunetlearningdense}, have been commonly used and demonstrate strong performance in this domain.


However, CNN-based methods require large, annotated datasets and significant computational resources. Their success hinges on diverse and extensive training datasets tailored to various muscle morphologies, which imposes high computational overhead and increases the dependency on manual annotations \cite{hesamian2019deep}.

Another key limitation of CNN-based approaches is their subject-specific nature. Models trained on specific demographics, such as pediatric datasets, often struggle to generalize to adults or females due to physiological variations \cite{lin2024automatic}.

In addition, the segmentation results produced by deep learning models often lack explainability \cite{reyes2020interpretability, salahuddin2022transparency}. Although CNN-based models can make reasonably accurate predictions through their training processes, when the boundaries are not discernible to the human eye, the segmentation results from CNN models still lack interpretability. 

These limitations, i.e., data dependency, computational cost, lack of generalizability, and limited interpretability, underscore the need for more adaptable, efficient, and explainable muscle segmentation approaches.

\textit{Feature Extraction}

Feature extraction methods can be broadly categorized into corner-based and structural approaches. Corner-based detectors, such as the Harris corner detector \cite{harris1988combined} and the FAST \cite{rosten2006machine}, are designed to identify localized points exhibiting significant intensity changes in multiple directions. While effective for tracking salient points in general scenes \cite{zhang2025plkcalib}, these methods are limited in their ability to capture continuous edge structures, which is an essential characteristic for delineating anatomical boundaries in medical imaging.

Structural approaches, such as those based on wavelet transformation \cite{mallat1989theory}, offer a more comprehensive means of detecting such boundaries. 2D discrete wavelet transform (DWT) decomposes images into hierarchical representations across spatial scales, isolating both coarse approximations and fine detail components \cite{brannock2006edge}. This enables the accurate localization of intensity transitions corresponding to edges of varying orientations and scales. Compared to traditional edge detection algorithms, wavelet-based methods have demonstrated superior accuracy in identifying continuous anatomical contours \cite{brannock2006edge}. Their sensitivity to gradient changes at multiple scales and orientations makes them particularly well suited for tasks where structural continuity is critical. 

\textit{Feature Tracking}

Descriptor-based tracking methods fundamentally rely on feature extraction and feature matching. These techniques capture local visual features using handcrafted \cite{park2022eigencontours} or learned descriptors \cite{zhang20213d, zhang2025nerfvio}, and establish correspondences across frames based on descriptor similarity. While effective in general-purpose tracking, these methods often struggle in medical imaging scenarios due to limited texture variation and high pixel similarity, which hinder the extraction and matching of distinctive local features.


Direct tracking methods, such as the Lucas-Kanade optical flow method \cite{lucas1981iterative}, offer an alternative by estimating motion based on image gradients and leveraging the global context. These methods have been successfully utilized in diverse fields, including motion estimation \cite{Boretti2022icra}, real-time video processing \cite{plyer2016massively}, and precipitation forecasting \cite{liu2015new}. Nevertheless, their application in medical imaging, particularly for precise keypoint tracking tasks, remains relatively underexplored, suggesting a promising direction for further research.


\section*{Methods}

This section introduces the dataset preparation and outlines the proposed method, including the keypoint selection and tracking algorithm.

\textit{Data Preparation}

The dataset utilized in this study consists of lower-body MRI scan from 10 typically developing (TD) children, identified as TD01 through TD10. Each subject undergoes imaging during two separate sessions (S1 and S2) as outlined in Table \ref{table:datasetOverview}. In total, the dataset includes 11,164 PNG images for analysis.

\begin{table}[t]
    \centering
    \caption{Datasets Information.}
    \vspace{0.3cm}
    \renewcommand{\arraystretch}{1.25}
    \begin{tabular}{|c|c|c|c|}
    \hline
    \textbf{Dataset}      & \textbf{Axial} & \textbf{Dataset}      & \textbf{Axial} \\ \hline
    \fontsize{11pt}{11pt}\selectfont
    TD01\_S1\_MRI & 585 & \fontsize{11pt}{11pt}\selectfont TD01\_S2\_MRI & 514\\ \hline
    \fontsize{11pt}{11pt}\selectfont
    TD02\_S1\_MRI & 570 &
    \fontsize{11pt}{11pt}\selectfont
    TD03\_S1\_MRI & 572 \\ \hline
    \fontsize{11pt}{11pt}\selectfont
    TD04\_S1\_MRI & 547 & \fontsize{11pt}{11pt}\selectfont TD04\_S2\_MRI & 930  \\ \hline
    \fontsize{11pt}{11pt}\selectfont
    TD05\_S1\_MRI & 549 & \fontsize{11pt}{11pt}\selectfont TD05\_S2\_MRI & 1,042 \\ \hline
    \fontsize{11pt}{11pt}\selectfont
    TD06\_S1\_MRI & 544 & \fontsize{11pt}{11pt}\selectfont TD06\_S2\_MRI & 1,042  \\ \hline
    \fontsize{11pt}{11pt}\selectfont
    TD07\_S1\_MRI & 544 & \fontsize{11pt}{11pt}\selectfont TD07\_S2\_MRI & 1,042 \\ \hline
    \fontsize{11pt}{11pt}\selectfont
    TD08\_S1\_MRI & 545 & \fontsize{11pt}{11pt}\selectfont TD08\_S2\_MRI & 1,042  \\ \hline
    \fontsize{11pt}{11pt}\selectfont
    TD09\_S1\_MRI & 547 &
    \fontsize{11pt}{11pt}\selectfont
    TD10\_S1\_MRI & 549 \\ \hline
    \textbf{Total} & \multicolumn{3}{c|}{\textbf{11,164}} \\ \hline
    \end{tabular}
    \label{table:datasetOverview}
\end{table}

\begin{figure}[tb]
    \centering
    \includegraphics[width=0.50\textwidth]{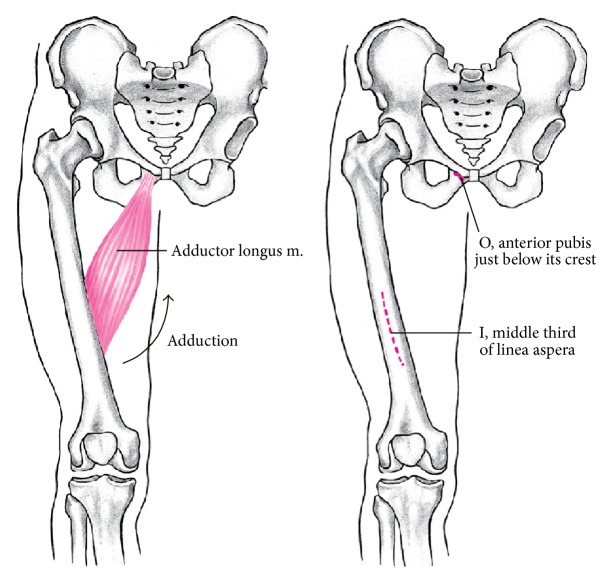}
    \caption{Adductor longus muscle \cite{van2015rare}.}
    \label{fig: adductor_figure}
\end{figure}

Graduate students are hired to generate the ground truth annotations for the adductor longus muscle (shown in Figure \ref{fig: adductor_figure}) using the semi-automated segmentation tool LabelMe \cite{Wada_Labelme_Image_Polygonal}. An experienced radiation oncologist subsequently reviews and verifies these annotations to ensure accuracy and reliability.

\textit{Keypoint Selection and Tracking}

The implementation of tracking methods for muscle boundary analysis leverages the minimal morphological changes observed across consecutive MRI slices in axial, coronal, or sagittal view. These slices, typically spaced 1 mm apart, exhibit significant continuity in muscle shape, making tracking algorithms particularly well-suited for accurately following keypoints along muscle boundaries.

\begin{figure}[t]
    \centering
    \includegraphics[width=0.95\textwidth]{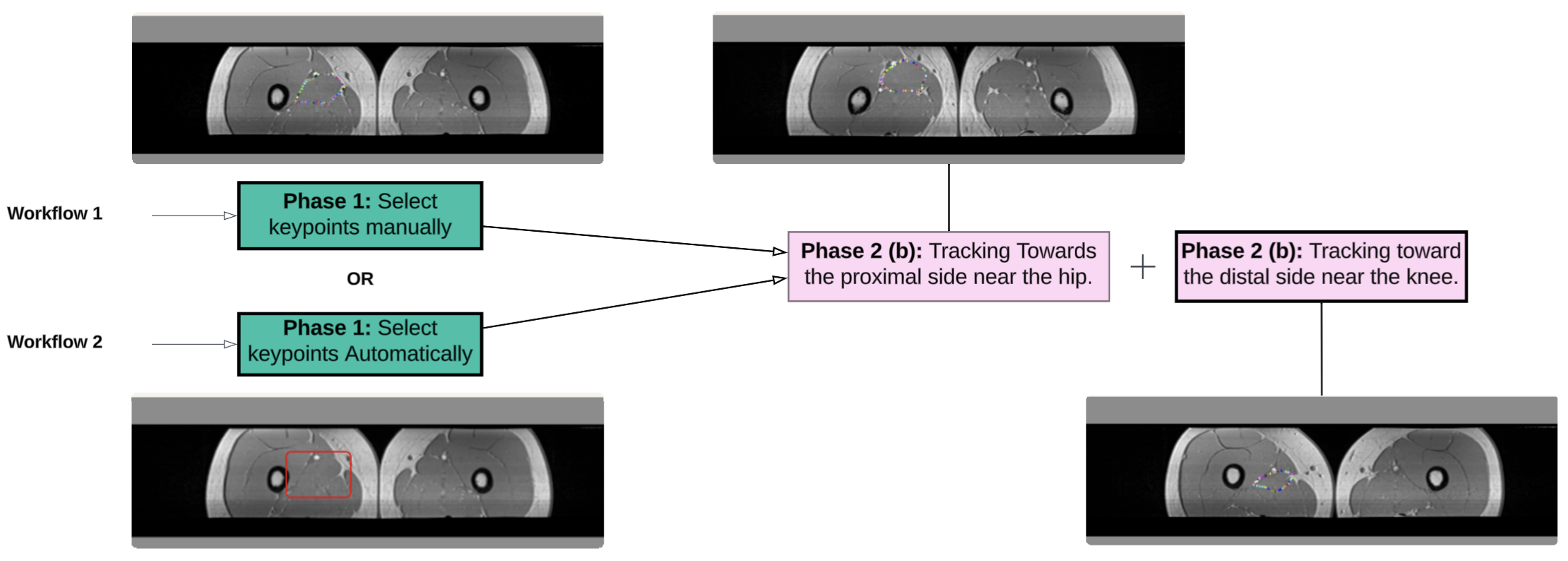}
    \caption{The workflow of the tracking process.}
    \label{fig: workflow}
\end{figure}

The muscle tracking workflow comprises two main phases: (1) keypoint selection, and (2) feature tracking, as illustrated in Figure \ref{fig: workflow}. In Phase 1, keypoints are selected either manually or automatically. In the manual method, i.e., Workflow 1 in Figure \ref{fig: workflow}, approximately 40 keypoints are visually identified on the central frame of the target muscle region by analyzing anatomical boundaries. The automatic method, i.e., Workflow 2 in Figure \ref{fig: workflow}, selects keypoints within a user-defined region of interest (ROI) based on frequency characteristics.

\begin{figure}[t]
    \centering
    \includegraphics[width=0.95\textwidth]{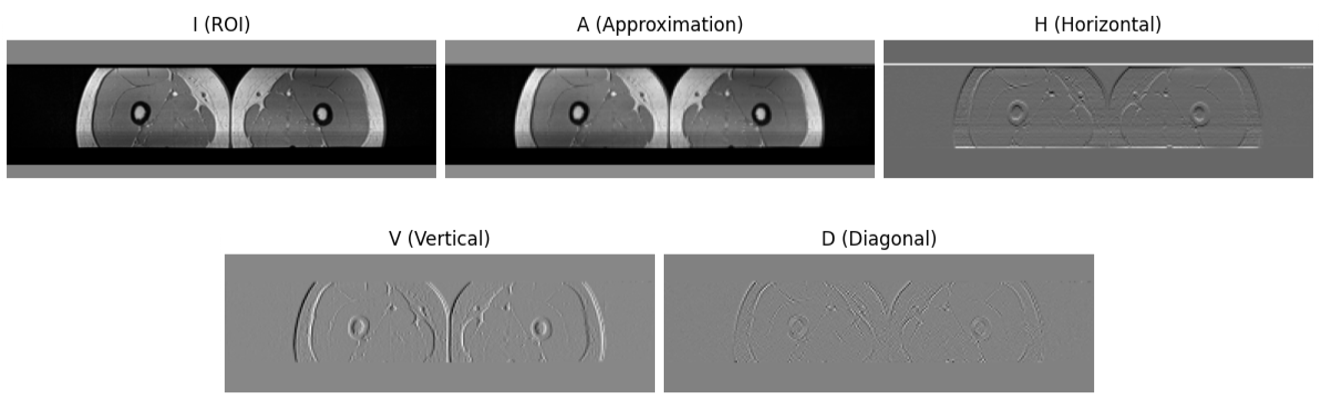}
    \caption{2D Haar DWT example.}
    \label{fig: haar}
\end{figure}

Specifically, in Workflow 2, 2D Haar DWT \cite{mallat1989theory} is applied to the grayscale ROI highlighted by the red bounding box to detect areas of high-frequency content, which often correspond to structural boundaries within anatomical structures. Mathematically, the 2D Haar DWT decomposes the ROI into one approximation component and three detail components as in equation \eqref{eq:DWT}:
\begin{equation}
{DWT}_{{Haar}}(I) \rightarrow (A, H, V, D)
\label{eq:DWT}
\end{equation}
, where $I$ is the grayscale ROI, $A$ is the approximation (low-frequency) component, and $H$, $V$, and $D$ are the horizontal, vertical, and diagonal detail components, respectively. An example of 2D Haar DWT on lower body MRI is shown in Figure \ref{fig: haar}. A magnitude map $M(x, y)$ is then computed by utilizing equation \eqref{eq:mag}:
\begin{equation}
M(x, y) = |H(x, y)| + |V(x, y)| + |D(x, y)|
\label{eq:mag}
\end{equation}
, where $x$ and $y$ denote the spatial coordinates of each pixel. Keypoints are extracted and represented as a set of points $\mathcal{S}_{KP}$, defined by locations where the magnitude map exceeds a predefined threshold $t$:
\begin{equation}
\mathcal{S}_{KP} = \{(x, y) \mid M(x, y) > t\}
\end{equation}
These keypoints are then rescaled to match the original image resolution.

In the feature tracking phase, i.e., Phase 2, the selected keypoints are propagated across adjacent frames using the Lucas-Kanade optical flow method. Tracking is conducted in both upward and downward directions. Phase 2 (a) shows the upward direction toward the proximal side near the hip, while Phase 2 (b) shows the downward direction toward the distal side near the knee.

\textit{Segmentation Mask Generation}

To generate the segmentation mask, a convex hull algorithm is employed on a set of points corresponding to the tracked features. The convex hull represents the smallest convex polygon that encompasses all the given points, effectively mitigating irregularities and producing a tightly bound, and closed contour. This method ensures that the generated mask dynamically adapts to the convex boundary of the tracked features in each frame, enhancing its robustness and suitability for applications such as object tracking and motion analysis.

\textit{Performance Metrics}

For evaluation, we employ the DSC to quantify the pixel-wise concordance between a predicted segmentation and the corresponding ground truth. The coefficient is calculated as follows in equation \eqref{eq:dice}: 

\begin{equation}
    \text{DSC}\left(A, B\right) = \frac{2 |A \cap B|}{|A| + |B|}
    \label{eq:dice}
\end{equation}

\noindent
, where \( A \) represents the predicted set of pixels and \( B \) denotes the ground truth. The DSC ranges from 0, indicating no overlap, to 1, signifying perfect overlap. 


\section*{Preliminary Experimental Results}

\begin{figure}[t]
    \centering
    \includegraphics[width=1.00\textwidth]{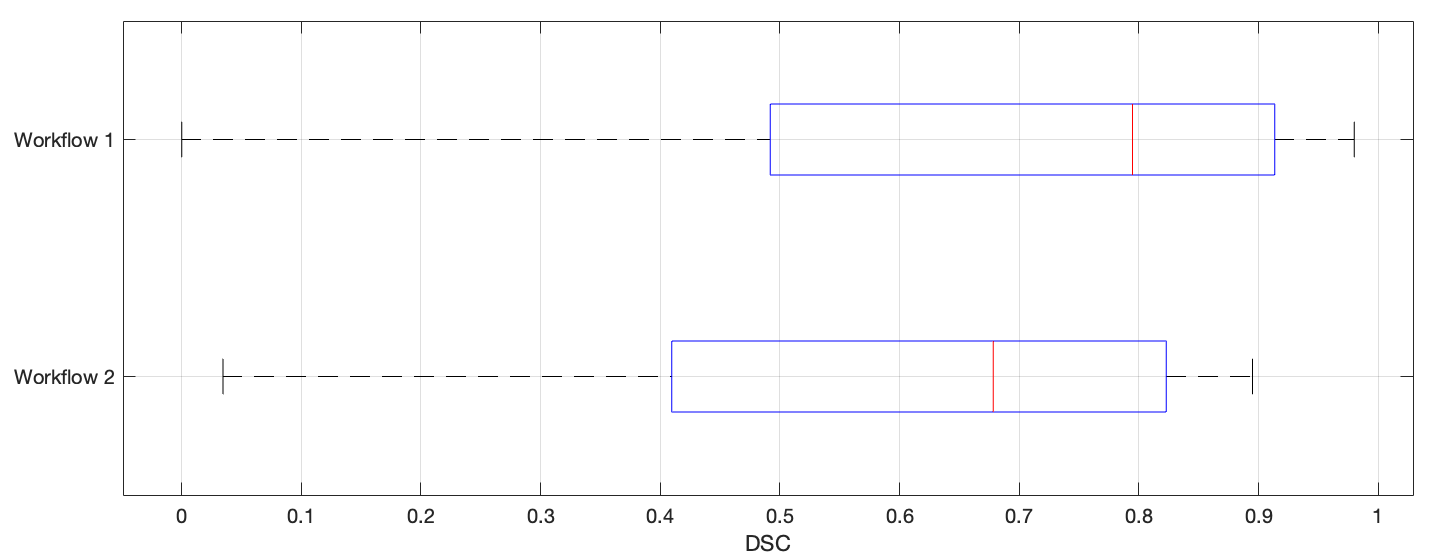}
    \caption{Distribution of DSC for two Workflows.}
    \label{fig: boxplot}
\end{figure}

\begin{figure}[t]
    \centering
    \includegraphics[width=0.95\textwidth]{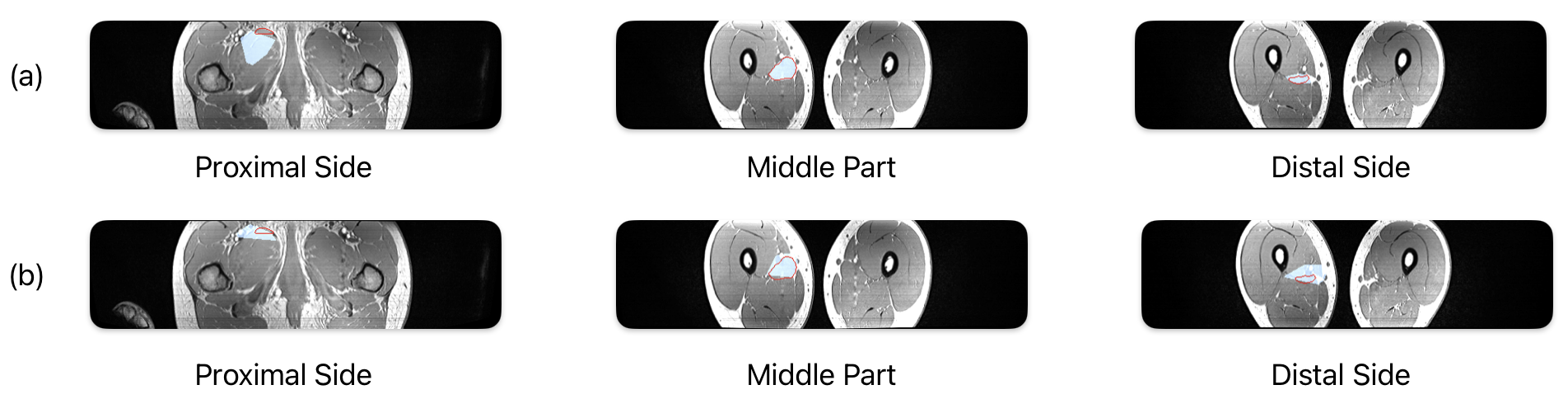}
    \caption{The figure presents segmentation results, with row (a) showing predictions from the Workflow 1 and row (b) displaying results from the Workflow 2. Predicted segmentations are marked in blue, while red boundaries denote ground-truth annotations. For Workflow 1, the DSCs are 0.015, 0.962, and 0.709 from left to right. For Workflow 2, the corresponding DSCs are 0.263, 0.772, and 0.305.}
    \label{fig: track_example}
\end{figure}

\begin{figure}[t]
    \centering
    \includegraphics[width=0.95\textwidth]{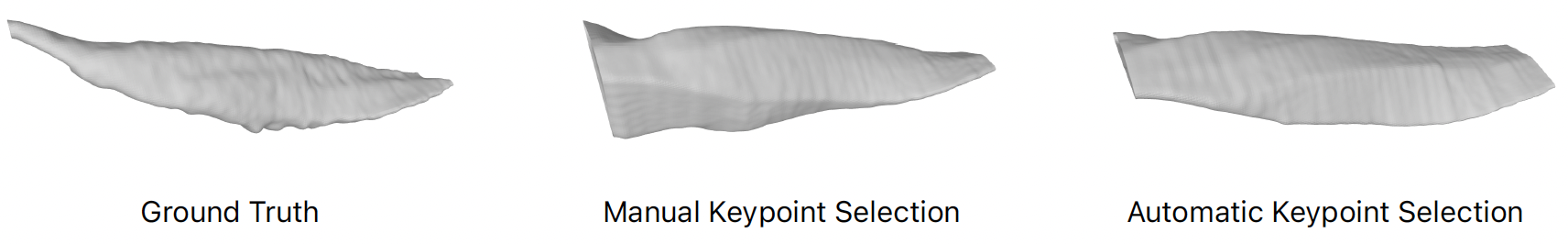}
    \caption{This figure shows the 3D reconstruction results. From left to right are the ground truth, the output from Workflow 1, and the output from Workflow 2.}
    \label{fig: 3d}
\end{figure}

The proposed method achieved a DSC of 0.670 ± 0.300 for tracking the adductor longus muscle with manual initial keypoint selection. In contrast, the method achieved a DSC of 0.604 ± 0.250 with automatic initial keypoint selection using 2D Haar DWT. The corresponding box plot is presented in Figure \ref{fig: boxplot}. Representative segmentation results for different partitions of the adductor longus muscle, i.e., the proximal, middle, and distal regions, are illustrated in Figure \ref{fig: track_example}. The 3D reconstruction results for both Workflows are shown in Figure \ref{fig: 3d}.

Although Workflow 2 demonstrates slightly lower mean and median accuracy compared to Workflow 1, it exhibits a smaller standard deviation and no instances of a 0 DSC indicating greater stability and consistency when keypoints are selected automatically. Segmentation outputs further reveal that Workflow 2 more reliably preserves the position of the target muscle throughout the sequence with reduced keypoint drift. In contrast, keypoints in Workflow 1 tend to drift significantly, particularly when tracking toward the proximal region, likely due to the lower robustness of manually selected keypoints. Nevertheless, Workflow 1 achieves more accurate segmentation in the middle and distal regions of the muscle. This difference is attributed to the manually selected keypoints being more precisely contouring the target muscle boundaries, whereas the automated method inadvertently includes the keypoints of the boundaries of adjacent muscles, resulting in slight over-segmentation during tracking.

Overall, both Workflows display a relatively wide interquartile range (IQR), reflecting some variability in performance. However, their average DSC remains within the range of 0.60 to 0.70. This performance is comparable to that of state-of-the-art CNN-based segmentation models \cite{ni2019automatic}.

\section*{Conclusion}

Previous research in muscle segmentation has primarily relied on CNNs, which have shown strong performance but are limited by their need for large annotated datasets, high computational costs, and poor interpretability.

In this study, we propose a training-free segmentation framework that combines keypoint selection with Lucas-Kanade optical flow tracking to delineate muscle boundaries in MRI scans. Our method supports both manual and automatic keypoint selection strategies, the latter using wavelet-based frequency analysis to identify salient anatomical features. The resulting keypoints are propagated across frames, and segmentation masks are generated using a convex hull algorithm. This approach achieves competitive accuracy, with average DSC ranging from 0.60 to 0.70, while significantly reducing computational cost. By explicitly tracking anatomical landmarks and eliminating the need for model training, our method offers a scalable, efficient, and explainable alternative to conventional deep learning-based approaches, with broad applicability across diverse subject populations.

\section*{Future Work}

While our current approach demonstrates promising results, further validation across a broader range of muscles is necessary to ensure its generalizability and robustness. Specifically, this requires access to a more extensive dataset with annotated muscle boundaries across different anatomical regions and subject populations. The inclusion of diverse muscle datasets will allow us to refine our methodology and address potential variations in muscle morphology.

Currently, keypoints are identified only once at the beginning of the tracking process. Inspired by the work of simultaneous localization and mapping (SLAM) \cite{Zhang2024ccta}, where proprioceptive input predicts pose and exteroceptive sensors refine the estimate based on environmental constraints, future iterations of this work will incorporate additional inputs during tracking, such as reselecting keypoints or adjusting the bounding box, to better guide the tracking process. This strategy offers a promising means of mitigating tracking-induced errors.

Moreover, a significant portion of volumetric medical data, including MRI, is anisotropic, as the through-plane resolution is often much lower than the in-plane resolution \cite{mahesh2013essential}. Drawing inspiration from techniques such as 2.5D Cross-Slice Attention \cite{hung2024csam} and crowdsourced map updating in distributed systems \cite{zhang2024distributed}, motion patterns from adjacent acquired MRI scans could be leveraged to predict the position of muscle contours in new data. This temporal or inter-scan context could improve prediction accuracy, especially for muscles with predictable movement or deformation patterns. 

Together, these enhancements hold promise for improving both the reliability and precision of the keypoint-tracking-based segmentation framework in medical imaging applications.

Looking ahead, we also intend to integrate the finalized tracking algorithm into MedVis Suite \cite{Liu2025}, an educational platform designed to support learning in computer vision for medical imaging. Incorporating this algorithm into MedVis Suite will enable students to visually explore and interact with muscle segmentation processes, including keypoint selection, boundary tracking, and 3D reconstruction. This hands-on approach aims to foster a deeper, more intuitive understanding of the algorithm’s functionality and its practical application in real-world medical imaging scenarios.


\section*{Acknowledgment}
This project was funded in part by the Northeastern TIER 1 seed grant.



\vspace{4\baselineskip}\vspace{-\parskip} 
\footnotesize 
\bibliographystyle{ieeetr} 
\bibliography{references.bib}

\begin{thebibliography}{10}

\bibitem{zhu2021deep}
J.~Zhu, B.~Bolsterlee, B.~V. Chow, C.~Cai, R.~D. Herbert, Y.~Song, and
  E.~Meijering, ``Deep learning methods for automatic segmentation of lower leg
  muscles and bones from mri scans of children with and without cerebral
  palsy,'' {\em NMR in Biomedicine}, vol.~34, no.~12, p.~e4609, 2021.

\bibitem{ni2019automatic}
R.~Ni, C.~H. Meyer, S.~S. Blemker, J.~M. Hart, and X.~Feng, ``Automatic
  segmentation of all lower limb muscles from high-resolution magnetic
  resonance imaging using a cascaded three-dimensional deep convolutional
  neural network,'' {\em Journal of Medical Imaging}, vol.~6, no.~4,
  pp.~044009--044009, 2019.

\bibitem{ketkar2021convolutional}
N.~Ketkar, J.~Moolayil, N.~Ketkar, and J.~Moolayil, ``Convolutional neural
  networks,'' {\em Deep learning with Python: learn best practices of deep
  learning models with PyTorch}, pp.~197--242, 2021.

\bibitem{Dice1945measures}
L.~R. Dice, ``Measures of the amount of ecologic association between species,''
  {\em Ecology}, vol.~26, no.~3, pp.~297--302, 1945.

\bibitem{hesamian2019deep}
M.~H. Hesamian, W.~Jia, X.~He, and P.~Kennedy, ``Deep learning techniques for
  medical image segmentation: achievements and challenges,'' {\em Journal of
  digital imaging}, vol.~32, pp.~582--596, 2019.

\bibitem{reyes2020interpretability}
M.~Reyes, R.~Meier, S.~Pereira, C.~A. Silva, F.-M. Dahlweid, H.~v.
  Tengg-Kobligk, R.~M. Summers, and R.~Wiest, ``On the interpretability of
  artificial intelligence in radiology: challenges and opportunities,'' {\em
  Radiology: artificial intelligence}, vol.~2, no.~3, p.~e190043, 2020.

\bibitem{salahuddin2022transparency}
Z.~Salahuddin, H.~C. Woodruff, A.~Chatterjee, and P.~Lambin, ``Transparency of
  deep neural networks for medical image analysis: A review of interpretability
  methods,'' {\em Computers in biology and medicine}, vol.~140, p.~105111,
  2022.

\bibitem{Ronneberger2015u}
O.~Ronneberger, P.~Fischer, and T.~Brox, ``U-net: Convolutional networks for
  biomedical image segmentation,'' in {\em Medical image computing and
  computer-assisted intervention--MICCAI 2015: 18th international conference,
  Munich, Germany, October 5-9, 2015, proceedings, part III 18}, pp.~234--241,
  Springer, 2015.

\bibitem{3dunetlearningdense}
Özgün Çiçek, A.~Abdulkadir, S.~S. Lienkamp, T.~Brox, and O.~Ronneberger,
  ``3d u-net: Learning dense volumetric segmentation from sparse annotation,''
  2016.

\bibitem{lin2024automatic}
Z.~Lin, W.~H. Henson, L.~Dowling, J.~Walsh, E.~Dall’Ara, and L.~Guo,
  ``Automatic segmentation of skeletal muscles from mr images using modified
  u-net and a novel data augmentation approach,'' {\em Frontiers in
  Bioengineering and Biotechnology}, vol.~12, p.~1355735, 2024.

\bibitem{harris1988combined}
C.~Harris, M.~Stephens, {\em et~al.}, ``A combined corner and edge detector,''
  in {\em Alvey vision conference}, vol.~15, pp.~10--5244, Citeseer, 1988.

\bibitem{rosten2006machine}
E.~Rosten and T.~Drummond, ``Machine learning for high-speed corner
  detection,'' in {\em Computer Vision--ECCV 2006: 9th European Conference on
  Computer Vision, Graz, Austria, May 7-13, 2006. Proceedings, Part I 9},
  pp.~430--443, Springer, 2006.

\bibitem{zhang2025plkcalib}
Y.~Zhang, J.~Xu, and W.~Ren, ``Plk-calib: Single-shot and target-less
  lidar-camera extrinsic calibration using pl$\backslash$" ucker lines,'' {\em
  arXiv preprint arXiv:2503.07955}, 2025.

\bibitem{mallat1989theory}
S.~G. Mallat, ``A theory for multiresolution signal decomposition: the wavelet
  representation,'' {\em IEEE transactions on pattern analysis and machine
  intelligence}, vol.~11, no.~7, pp.~674--693, 1989.

\bibitem{brannock2006edge}
E.~Brannock and M.~Weeks, ``Edge detection using wavelets,'' in {\em
  Proceedings of the 44th annual ACM Southeast Conference}, pp.~649--654, 2006.

\bibitem{park2022eigencontours}
W.~Park, D.~Jin, and C.-S. Kim, ``Eigencontours: Novel contour descriptors
  based on low-rank approximation,'' in {\em Proceedings of the IEEE/CVF
  Conference on Computer Vision and Pattern Recognition}, pp.~2667--2675, 2022.

\bibitem{zhang20213d}
Y.~Zhang, J.~Song, and S.~Li, ``3d object detection and tracking using
  monocular camera in carla,'' in {\em 2021 IEEE International Conference on
  Electro Information Technology (EIT)}, pp.~067--072, IEEE, 2021.

\bibitem{zhang2025nerfvio}
Y.~Zhang, D.~Wang, J.~Xu, M.~Liu, P.~Zhu, and W.~Ren, ``Nerf-vio: Map-based
  visual-inertial odometry with initialization leveraging neural radiance
  fields,'' {\em arXiv preprint arXiv:2503.07952}, 2025.

\bibitem{lucas1981iterative}
B.~D. Lucas and T.~Kanade, ``An iterative image registration technique with an
  application to stereo vision,'' in {\em IJCAI'81: 7th international joint
  conference on Artificial intelligence}, vol.~2, pp.~674--679, 1981.

\bibitem{Boretti2022icra}
C.~Boretti, P.~Bich, Y.~Zhang, and J.~Baillieul, ``Visual navigation using
  sparse optical flow and time-to-transit,'' in {\em 2022 International
  Conference on Robotics and Automation (ICRA)}, pp.~9397--9403, 2022.

\bibitem{plyer2016massively}
A.~Plyer, G.~Le~Besnerais, and F.~Champagnat, ``Massively parallel lucas kanade
  optical flow for real-time video processing applications,'' {\em Journal of
  Real-Time Image Processing}, vol.~11, pp.~713--730, 2016.

\bibitem{liu2015new}
Y.~Liu, D.-G. Xi, Z.-L. Li, and Y.~Hong, ``A new methodology for
  pixel-quantitative precipitation nowcasting using a pyramid lucas kanade
  optical flow approach,'' {\em Journal of Hydrology}, vol.~529, pp.~354--364,
  2015.

\bibitem{Wada_Labelme_Image_Polygonal}
K.~Wada, ``{Labelme: Image Polygonal Annotation with Python}.''

\bibitem{van2015rare}
R.~Van De~Kimmenade, C.~Van~Bergen, P.~Van~Deurzen, and R.~Verhagen, ``A rare
  case of adductor longus muscle rupture,'' {\em Case reports in orthopedics},
  vol.~2015, no.~1, p.~840540, 2015.

\bibitem{Zhang2024ccta}
Y.~Zhang, P.~Zhu, and W.~Ren, ``Pl-cvio: Point-line cooperative visual-inertial
  odometry,'' in {\em 2023 IEEE Conference on Control Technology and
  Applications (CCTA)}, pp.~859--865, 2023.

\bibitem{mahesh2013essential}
M.~Mahesh, ``The essential physics of medical imaging,'' {\em Medical physics},
  vol.~40, no.~7, p.~077301, 2013.

\bibitem{hung2024csam}
A.~L.~Y. Hung, H.~Zheng, K.~Zhao, X.~Du, K.~Pang, Q.~Miao, S.~S. Raman,
  D.~Terzopoulos, and K.~Sung, ``Csam: A 2.5 d cross-slice attention module for
  anisotropic volumetric medical image segmentation,'' in {\em Proceedings of
  the IEEE/CVF Winter Conference on Applications of Computer Vision},
  pp.~5923--5932, 2024.

\bibitem{zhang2024distributed}
Y.~Zhang, M.~Greiff, W.~Ren, and K.~Berntorp, ``Distributed road-map monitoring
  using onboard sensors,'' in {\em 2024 American Control Conference (ACC)},
  pp.~5049--5054, IEEE, 2024.

\bibitem{Liu2025}
M.~Liu, D.~Zhang, Y.~Chen, T.~Gong, H.~Kainz, S.~Song, and J.~Lee, ``Medvis
  suite: A framework for mri visualization and u-net-based bone segmentation
  with in-depth evaluation,'' in {\em BIO Web of Conferences}, vol.~163,
  p.~04001, EDP Sciences, 2025.

\end{thebibliography}

\end{document}